\documentclass[runningheads]{llncs}
\usepackage{graphicx}
\usepackage{comment}
\usepackage{amsmath,amssymb} 
\usepackage{color}

\usepackage{microtype}
\usepackage{subfigure}
\usepackage{booktabs} 

\usepackage{hyperref}

\usepackage{lineno}
\usepackage{times}
\usepackage{url}

\usepackage{algorithm}
\usepackage{algorithmic}

\usepackage{multirow}
\usepackage{placeins}

\begin{document}
\pagestyle{headings}
\mainmatter
\def\ECCVSubNumber{7302}  

\title{Fixed-size Objects Encoding for Visual Relationship Detection} 

\author{Hengyue Pan, Xin Niu, Rongchun Li, Siqi Shen, Yong Dou}
\institute{PDL Lab, School of Computer, National University of Defense Technology}

\maketitle

\begin{abstract}
In this paper, we propose a fixed-size object encoding method (FOE-VRD) to improve performance of visual relationship detection tasks. Comparing with previous methods, FOE-VRD has an important feature, i.e., it uses one fixed-size vector to encoding all objects in each input image to assist the process of relationship detection. Firstly, we use a regular convolution neural network as a feature extractor to generate high-level features of input images. Then, for each relationship triplet in input images, i.e., $<$subject-predicate-object$>$, we apply ROI-pooling to get feature vectors of two regions on the feature maps that corresponding to bounding boxes of the subject and object. Besides the subject and object, our analysis implies that the results of predicate classification may also related to the rest objects in input images (we call them background objects). Due to the variable number of background objects in different images and computational costs, we cannot generate feature vectors for them one-by-one by using ROI pooling technique. Instead, we propose a novel method to encode all background objects in each image by using one fixed-size vector (i.e., FBE vector). By concatenating the 3 vectors we generate above, we successfully encode the objects using one fixed-size vector. The generated feature vector is then feed into a fully connected neural network to get predicate classification results. Experimental results on VRD database (entire set and zero-shot tests) show that the proposed method works well on both predicate classification and relationship detection. 
\dots
\end{abstract}

\section{Introduction}
\label{Sec_Intro}

To understand a given image, we should learn many information from it. One natural idea is that we need to know the locations and classes of foreground objects, and the corresponding task is called object detection in computer vision. In the past few years, many successful object detection algorithms have been proposed. \cite{viola2001robust} designed an object detection framework for face detection, which achieved high speed and good detection rate on hardware platforms with limited performance. With the development of hardware and related databases, such as ImageNet \cite{deng2009imagenet} and COCO \cite{lin2014microsoft}, increasing number of object detection algorithms tended to apply deep learning technique and considered to solve more general object detection tasks instead of only narrowed the application scope in face detection. SSD \cite{liu2016ssd} is a successful object detection algorithm, which can detect variety classes of objects. SSD removed proposal generation and feature resampling stages. Therefore, it simply used one network to do all computation, which made it easier to be trained and more user friendly in industrial practices. YOLO \cite{redmon2016you} is another very important object detection framework. Different from prior works, YOLO separated bounding boxes and corresponding class probabilities by defining object detection task as a regression problem. This operation resulted in a single detection network, which can be trained end-to-end. In \cite{girshick2015fast}, Fast RCNN was proposed for general object detection tasks. Based on R-CNN \cite{girshick2014rich} and SPPNets \cite{he2015spatial}, Fast R-CNN implemented single-stage training with multi-task objective function. Comparing with RCNN and SPPNets, Fast RCNN has faster training speed and higher detection quality. Faster RCNN \cite{ren2015faster} is an updated version of Fast RCNN, which introduced a Region Proposal Network (RPN) into the model to work with Fast RCNN network. RPNs and Fast R-CNN can be trained jointly, which obviously improve the efficiency of model learning. Experimental results proved that Faster R-CNN has good learning speed and state-of-the-art performance on many databases, and as a result, Faster R-CNN has been widely used in recent few years. Mask R-CNN \cite{he2017mask} is an extension of Faster R-CNN. Comparing with Faster R-CNN, Mask R-CNN introduced a new branch to predict an object mask. This branch should work together with the branch of bounding box recognition. In many object detection databases, Mask R-CNN achieved state-of-the-art performance. 

However, simply knowing the locations and class IDs of objects is far from enough. In many cases, two images that include the same kinds of objects may have quite different semantic meaning. This fact requires us to classify the relationship between objects in images. By doing this, we can build the logical connection between objects, which helps us to understand the images deeply and accurately. This task is the so-called Visual Relationship Detection. Recently, comparing with object detection, visual relationship detection attracts more attention, not only because it is much more challengeable then object detection, but also because it may help us to develop better algorithms to understand natural images automatically. Even though visual relationship detection is a relatively new research field, many excellent researches have been proposed in past few years. \cite{lu2016visual} proposed VRD algorithm, which was a successful research at the early stage of visual relationship detection. By combining a visual model and a language model, the VRD algorithm achieved good performance on several visual relationship detection databases. Moreover, this paper proposed to use Recall@X as the performance metric instead of mAP, and Recall@X is applied in many subsequent researches. In \cite{yu2017visual}, Yu et al. argued that it is unfair to simply evaluate the top detected relationship between each object pair since some correct predictions may be penalized mistakenly. Therefore, a new hyper-parameter $k$ was introduced, which is the number of the chosen predictions per object pair. The introducing of $k$ made the evaluation of visual relationship detection tasks more reasonable, thus most of novel works tend to apply $k$ in their experiments. In \cite{liang2018visual}, a visual relationship detection framework called Deep Structural Ranking was proposed. This method introduced structural ranking loss into the deep neural network framework to reduce the negative impact of incomplete annotations. Deep Structured Learning \cite{zhu2018deep} is another efficient relationship detection method. The deep structured model included feature-level and label-level relationship prediction, and the two prediction results should be weighted summarized for the final result. Moreover, this research presented to use SSVM loss as the optimization goal, which resulted in simpler and more independent optimization procedures. Zoom-Net \cite{yin2018zoom} considered to apply two pooling cells, i.e., Contrastive ROI Pooling and Pyramid ROI Pooling, to improve relationship detection ability of the network. The algorithm achieved good performance on several databases. 

In this paper, we propose a novel visual relationship detection algorithm called FOE-VRD. Comparing with the previous counterparts, the proposed method pay more attention to one problem: how to encode the objects in given images to improve the performance of visual relationship detection? The basis of this question is the consideration that the kind of relationship between two selected objects may not simply related to the two objects themselves, but all background objects in the image. Therefore, encoding all objects to a fixed-size vector may bring about conveniences to further computations. The main obstacle of this idea is fact that the number of objects in images are not fixed, thus it is not a trivial task to use a fixed-size vector to do object encoding. The main contribution of this paper is that we propose a novel method to generate a fixed-size feature vector to model all objects in one image, no matter how many objects are included in the image. The fixed-size feature vector discards most visual information, but keeps two important things: the class IDs and the relative distribution of background objects. By introducing the fixed-size background encoding (FBE) method, the proposed FOE-VRD algorithm achieves state-of-the-art performance on VRD database. 

\begin{figure}[t]
	\vskip 0.2in
	\begin{center}
		\centerline{\includegraphics[width=\columnwidth]{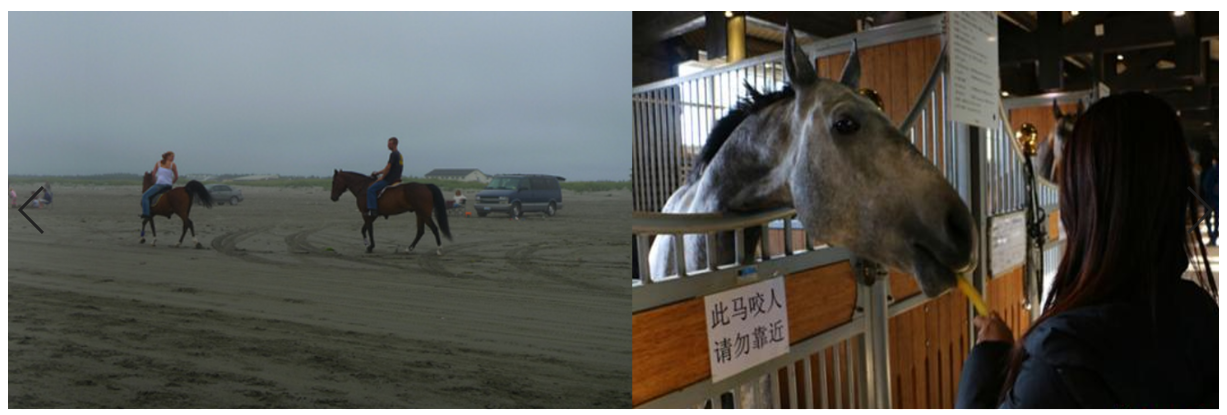}}
		\caption{$<$human-riding-horse$>$(left) and $<$human-feeding-horse$>$(right) may have quite different background objects. For example, in the images of $<$human-feeding-horse$>$, some items such as barriers and fodders may appear.}
		\label{ridingandfeeding}
	\end{center}
	\vskip -0.2in
\end{figure}

\section{Related Works}
\label{Sec_Related}
Generally speaking, the proposed FOE-VRD algorithm has some important bases, i.e., ROI pooling method, VRD algorithm, and FOFE algorithm. In this section, we review the core ideas of the three above mentioned methods. 

\subsection{ROI Pooling}

In \cite{girshick2015fast}, the ROI Pooling layer is proposed for object detection tasks. A ROI (Region Of Interest) is a rectangular region on feature maps of one convolution layer, which is corresponding to a foreground object of the input image. By introducing max-pooling operation, we can pool each ROI rectangle to a group of smaller fixed-size feature maps. By doing ROI Pooling, we can convert all ROIs to fixed-size features, despite the different size of their corresponding objects. This operation brings about convenience for further detection and classification operations. In this work, we apply ROI pooling to encode the subject and object of an given relationship triplet. 

\subsection{VRD Algorithm}

The VRD algorithm proposed in \cite{lu2016visual} is an important visual relationship detection algorithm, which serves as significant basis for many researches. The VRD algorithm introduced two models, i.e. visual model and language model, for  relationship detection tasks. Specifically, for each pair of objects in input images, a convolutional neural network served as the visual model to calculate the possibility of each kind of predicate. On the other hand, a projection function was trained to project all possible relationships into an embedding space, where the relationships with similar semantic meaning should close to each other. By proposing a objective function combined the loss of visual model and language model, the VRD model can be trained efficiently. This work reveals the importance of semantic information in visual relationship detection, which motivates us to consider more potential information to improve the performance of visual relationship detection.

\subsection{FOFE}

\begin{figure*}[ht]
	\vskip 0.2in
	\begin{center}
		\centerline{\includegraphics[width=\columnwidth]{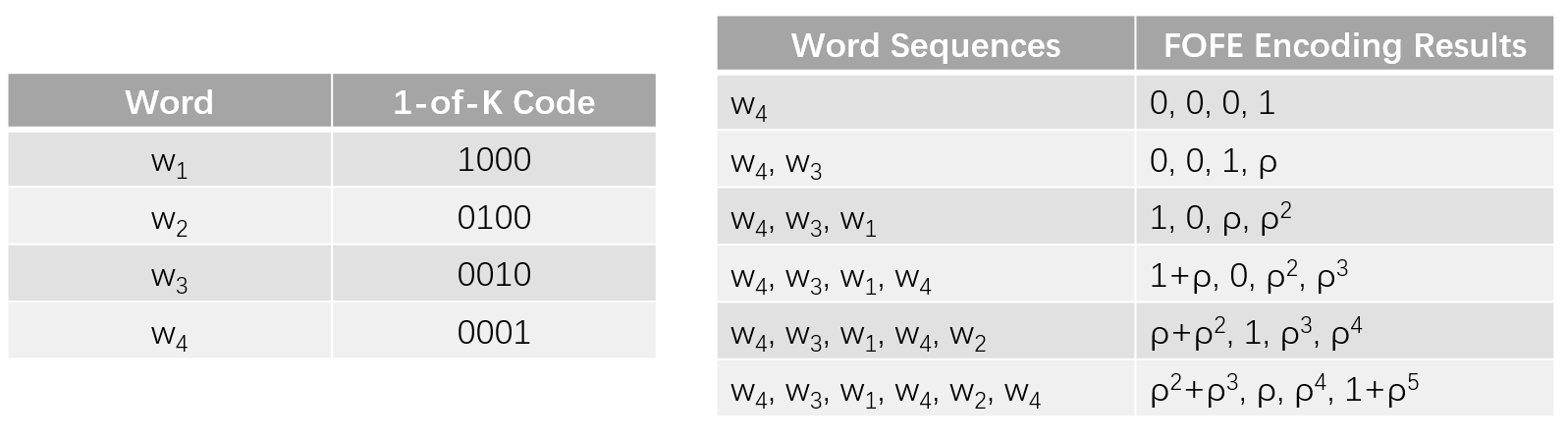}}
		\caption{The basic idea of FOFE algorithm. The left table includes the 1-of-K code of each word in the dictionary, and the right table shows an example of how to generate FOFE vector for a word sequence.}
		\label{FOFE_LM}
	\end{center}
	\vskip -0.2in
\end{figure*}

Fixed-size Ordinally-Forgetting Encoding (FOFE) \cite{zhang2015fixed} is an encoding method of natural language processing. The basic idea of FOFE is shown in Figure~\ref{FOFE_LM}: assuming that we have a dictionary that includes $K$ words (we use $K = 4$ as an example), it is obvious that we can use 1-of-K code to represent each word. For any word sequences $S = {w_1, w_2, ..., w_T}$, we can always use corresponding $K$-length vectors $V$ to represent them by introducing a forgetting factor $\rho$:

\begin{equation}
\label{FOFE_eq}
V_t = \rho V_{t-1} + c_t
\end{equation} 

where $c_t$ is the 1-of-K code of the word $w_t$, and $V_t$ is the FOFE vector of the sequence ${w_1, w_2, ..., w_t}$. 

Moreover, \cite{zhang2015fixed} proved the uniqueness of FOFE vector, which means that FOFE can serve as a good encoding method for language models. 

FOFE is an important basis of our work, since it provides a possible way to encode variable-length word sequences to fixed-size feature vectors. Based on FOFE, we propose Fixed-size Background Encoding (FBE) method to encode all objects in a given image to a fixed-size feature vector.

\section{Method}
\label{Sec_Method}

The most important motivation of our work is the consideration that the category of a predicate not only related to the corresponding subject and object, but also the other background objects in the whole image. Figure~\ref{ridingandfeeding} shows a good example: in both images the subjects are human beings and the objects are horses, but they have different predicates. In left image the predicate is riding, while in right image the predicate is feeding. The two images have quite different background objects. For instance, in the right image we can find barriers and fodders, which may not appear around the relationship triplet of $<$human-ridding-horse$>$. Therefore, finding a suitable way to model the background objects may bring about positive impacts on the performance of visual relationship detection. However, unlike subject and object, we cannot use ROI pooling to model all background objects. The first reason is related to the uncertain number of background objects, which means that we cannot use ROI pooling to generate a fixed-size feature vector for all background objects. This may disturb the further processing that applies fully-connected layers. The second reason is related to the large number of background objects. In variety of visual relationship detection databases, such as VRD \cite{lu2016visual} and VG \cite{krishnavisualgenome}, most images contain 10 to 20, or even more objects. Based on our network structure, if we use ROI pooling, each object will be converted to feature maps that contain $7*7*512 = 25088$ elements. This may result in very huge time and space complexity. Thus in practice, we tend to only use ROI pooling to process subjects and objects of relationship triplets, and try to find a better way to model all background objects. 

\begin{figure}[t]
	\vskip 0.2in
	\begin{center}
		\centerline{\includegraphics[width=\columnwidth]{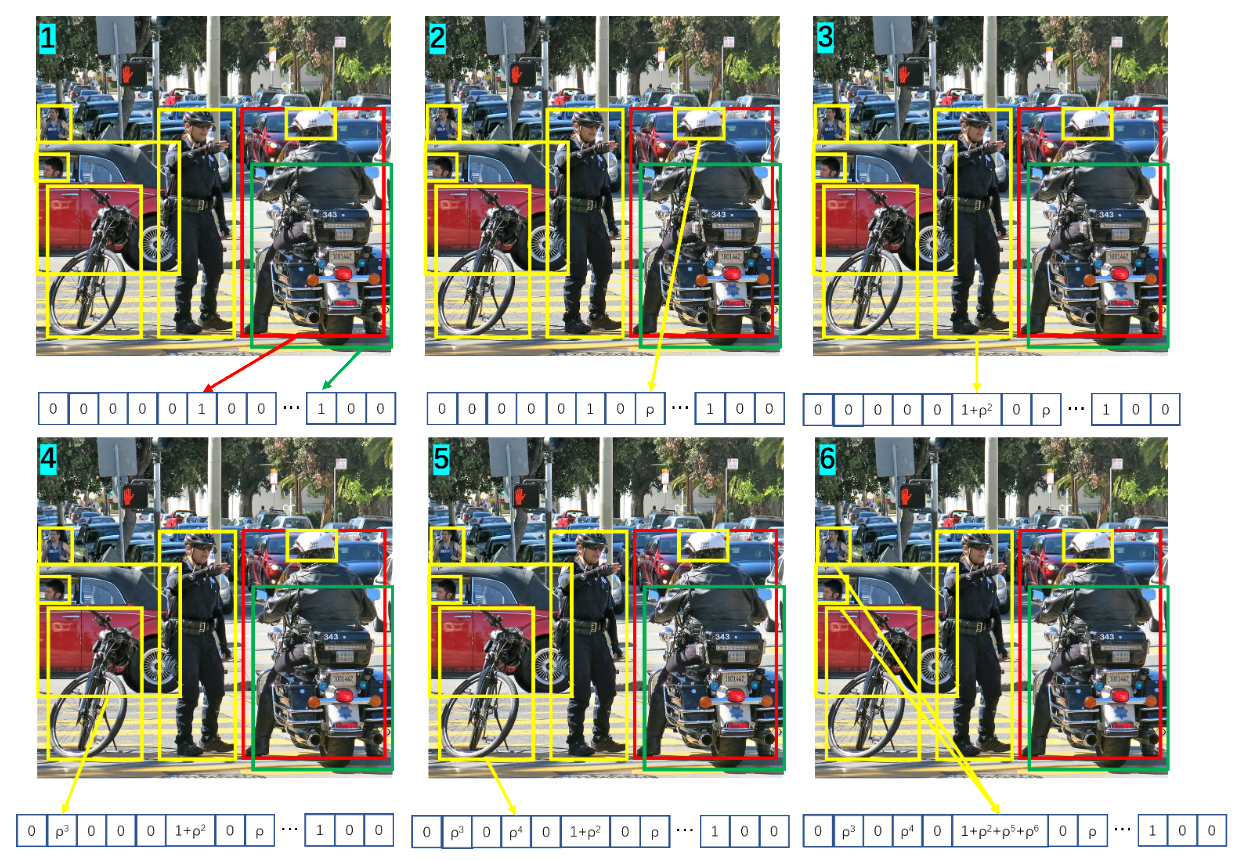}}
		\caption{The process to generate the FBE vector ${\bf B}$ of a given relationship triplet: (1) feed 1 to corresponding elements of the subject and object (here are person and motorbike respectively); (2) for the nearest background object to the subject and object, feed the forgetting factor $\rho$ to the corresponding element (here is helmet); (3) The second nearest background object is person again, so we add $\rho^2$ to the element of person; (4-6) we add all background elements one-by-one based on the distance from the center of the smallest bounding box that cover both subject and object, and finally we get ${\bf B}$ of the relationship $<$person-riding-motorbike$>$}
		\label{FOFE}
	\end{center}
	\vskip -0.2in
\end{figure}

\begin{figure*}[t]
	\vskip 0.2in
	\begin{center}
		\centerline{\includegraphics[width=\columnwidth]{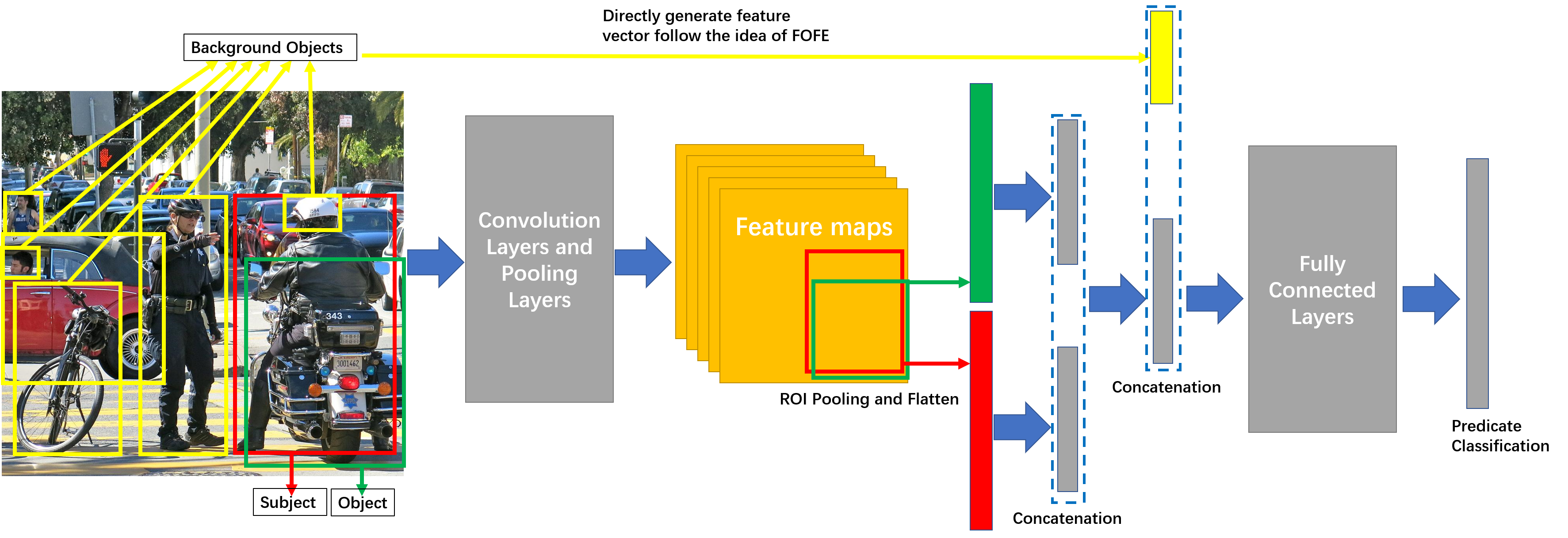}}
		\caption{The overall description of the proposed FOE-VRD algorithm}
		\label{foevrd-overall}
	\end{center}
	\vskip -0.2in
\end{figure*}

\subsection{Fixed-size Background Encoding}
\label{FBE}

Based on the FOFE algorithm we mentioned in Section~\ref{Sec_Related}, we propose Fixed-size Background Encoding (FBE) to model the background objects, and the algorithm outputs a fixed-size background vector (FBE vector) no matter how many background objects the input image contains. We use ${\bf B}$ to denote the FBE vector, and the length of ${\bf B}$ (denote by $L$) should equal to the number of object classes of the corresponding dataset (for instance, for VRD dataset $L = 100$, since the dataset contains $100$ categories of objects). We generate FBE vectors follow the rules below:

1. For the subject and object of one relationship triplet, we set the corresponding two elements (based on the class IDs of the subject ($c_s$) and object ($c_o$)) in ${\bf B}$ to $1$;

2. For every background object, we calculate the distance between the center of its bounding box and the center of the smallest bounding box that cover both of the subject and object. Assuming that we have $n$ background objects $b_1, b_2, ..., b_n$ with class IDs $c_1, c_2, ..., c_n$, and the corresponding distances are $d_1, d_2, ..., d_n$. Without loss of generality, we set $d_1 < d_2 < ... <d_n$.

3. We use $\rho$ to denote the forgetting factor ($ 0 < \rho < 1$). For $b_1$, we add $\rho$ to the $c_{1}$th element of ${\bf B}$, then for $b_2$, we add $\rho^2$ to the $c_{2}$th element. We repeat the operation above until all background objects have been processed. 

In Figure~\ref{FOFE}, we show an example of the generation of ${\bf B}$. Even though ${\bf B}$ discard most visual information, such as color and texture, of background objects, it still keeps two important information: the class IDs and the relative distribution of the background objects in the image. Here still leaves one question: is the FBE vector ${\bf B}$ unique? This question can be converted into Theorem~\ref{theorem-FBE} below:

\begin{theorem}[The uniqueness of FBE vectors]
	\label{theorem-FBE}
	 For two object lists with class ID sequences $C^1 = [c_s^1, c_o^1, c_1^1, c_2^1, ..., c_n^1]$ and $C^2 = [c_s^2, c_o^2, c_1^2, c_2^2, ..., c_n^2]$, we have ${\bf B}^1 \neq {\bf B}^2$ if ${\bf C^1} \neq {\bf C^2}$
	 
	\textbf{Proof:}
	Considering the $c_m^{1}$th element in ${\bf B}^1$. Assuming that we have a forgetting factor $\rho$, then we should add ${\rho}^m$ to ${\bf B}_{c_m^{1}}^1$ (if $m = s$ or $m = o$, we should add $1$ to ${\bf B}_{c_m^{1}}^1$). If $c_m^{2} \neq c_m^{1}$, we need to consider the inequality below to prove the uniqueness of FBE vectors:
	\begin{equation}
	\label{eq-summation}
	\sum_{i=m+1}^{\infty} \rho^i \ge \rho^m
	\end{equation}
	
	If the inequality in Eq.~\ref{eq-summation} is true, then it is possible that the FBE vectors are not unique. We use $S_{\infty}$ to denote the summation of the geometric series $\rho, \rho^2, \rho^3, ...$ ($ 0 < \rho < 1$), then we have:	
	\begin{equation}
	\label{eq-summation2}
	\sum_{i=m+1}^{\infty} \rho^i = S_{\infty} - S_{m} = \frac{\rho}{1-\rho} - \frac{\rho(1-\rho^m)}{1-\rho}  = \frac{\rho}{1-\rho} \cdot \rho^m
	\end{equation}
	
	${\bf (1)}$ If $0 < \rho < 0.5$, then $\rho$ always less then $1-\rho$, thus we have: 
	\begin{equation}
	\label{eq-summation3}
	\sum_{i=m+1}^{\infty} \rho^i < \rho^m
	\end{equation}
	
	therefore, in this case, FBE vectors are unique. 
	
	${\bf (2)}$ If $0.5 \le \rho < 1$, we can consider the situation that the number of terms (denote by $N$) of the left side in Eq.~\ref{eq-summation} is finite, and it is easy to show that $\sum_{i=m+1}^{N} \rho^i < \rho^m $ under the condition that $m > N+1$. Thus FBE vectors are unique in this situation.
	
	Even though we cannot strictly prove the uniqueness of FBE vectors when $0.5 \le \rho < 1$ and $m < N+1$, the situations that violate the uniqueness are extremely hard to happen in practice. We generated FBE vectors for all relationships in VRD database \cite{lu2016visual} with the forgetting factor $0.9$, and none of them violate the uniqueness theorem.  $\;\;\;\;\;  \blacksquare$
	
\end{theorem}

Based on the analysis above, the FBE vectors have potential to assist predicate classification. 

\subsection{The FOE-VRD Algorithm}

The proposed FOE-VRD model can be divided into 3 parts: 1. several convolution layers and corresponding pooling layers, which serve as feature extractor to generate abstract feature maps from input images; 2. relationship feature extractor, which is a fixed-size encoder to encoding each give relationship triplet; 3. several fully-connected layers to do the final predicate classification. 

By combining ROI pooling and FBE vectors,  we find a suitable way to encode all objects in an image. Specifically, for the subject and object in a relationship triplet, we use ROI pooling and a fully connected layer to generate their feature vectors (denote by ${\bf S}$ and ${\bf O}$ respectively). ${\bf S}$ and ${\bf O}$ are then concatenated, and feed into one fully connected layer to generate a feature vector $\bf M$. For the FBE vector ${\bf B}$, we use the method described in Section~\ref{FBE} to do the feature generation. Then $\bf M$ and ${\bf B}$ are concatenated as the feature vector (denote by ${\bf R}$) of the given relationship triplet. By using fully-connected layers to further extract features from ${\bf R}$, we finally get the results of predicate classification.

Figure~\ref{foevrd-overall} describes the overall structure of the proposed FOE-VRD algorithm. 

\begin{figure*}[t]
	\vskip 0.2in
	\begin{center}
		\centerline{\includegraphics[width=0.65\columnwidth]{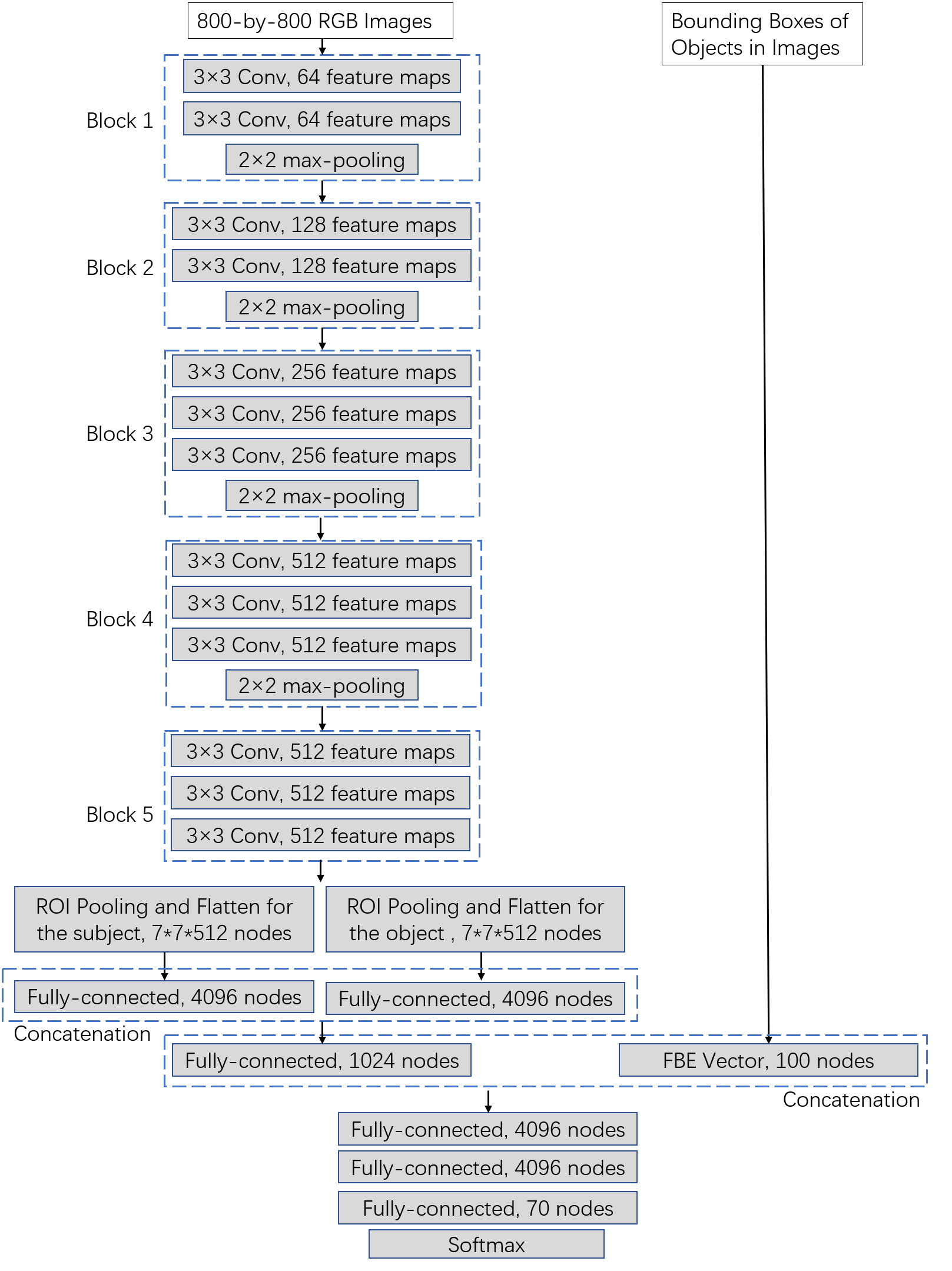}}
		\caption{The detailed network structure of FOE-VRD algorithm we used in our experiments}
		\label{foevrd-detail}
	\end{center}
	\vskip -0.2in
\end{figure*}

\section{Experimental Results}
\label{Sec_Experi}

In this section, we firstly present some important information of our experiments, which include database selection, experimental configurations and implementation details. Then we show the performance of FOE-VRD algorithm on predicate classification and relationship detection tasks. Moreover, we select many state-of-the-art relationship detection algorithms as baselines to compare with the proposed method. 

\subsection{Database}
We preform our experiments on Standford Visual Relationship Detection \cite{lu2016visual} database (VRD). VRD is an important database for visual relationship detection, which applied by many related works \cite{lu2016visual,mizzi2019optimising,jung2018visual,zhou2019visual,bin2019mr}. This database has 4000 training images and 1000 test images respectively. VRD database has 100 categories of objects and 70 categories of predicates, and the total number of the relationship triplets is 37993 (7701 types of triplets). Moreover, there are 1169 relationships (1029 types of triplets) only appear in the test set, thus we can use them to perform zero-shot tests. 

\subsection{Performance Measurements}
We consider two tasks to evaluate the proposed FOE-VRD algorithm, i.e., predicate classification and relationship detection. 

{\bf Predicate classification: } In case of predicate classification, the inputs of FOE-VRD include raw images, bounding boxes of objects in all images, and class IDs of all objects. Our algorithm predicts predicates between every pair of objects. This task shows the performance of FOE-VRD on the basis that the class IDs and locations of all objects are known. 

{\bf Relationship detection: } In case of relationship detection, the inputs of FOE-VRD only include raw images. We firstly use Faster RCNN algorithm \cite{ren2015faster} to do object detection on input images, then predict predicates between all pairs of detected objects by using FOE-VRD algorithm. The outputs are several relationship triplets $<$subject-predicate-object$>$. Comparing with predicate classification, relationship detection is more closed to real world applications, since in practice we cannot always have labeled images. 

For both tasks, we use Recall@100 as the performance measurement. Recall@X is widely used in visual relationship detection researches \cite{lu2016visual,zhu2018deep,liang2018visual,yin2018zoom}. To calculate Recall@X, we firstly collect in top X predictions in each image, then compute the average fraction of correct relationship among them \cite{lu2016visual}. 

\subsection{Implementation Details}

We implement our FOE-VRD algorithm on TensorFlow 1.6 \cite{tensorflow2015-whitepaper} platform. The network structure of FOE-VRD bases on VGG-16 network \cite{simonyan2014very} (see Fig.~\ref{foevrd-detail}) for more details). At the ROI pooling layer, we convert the regions of the subject and object to two vectors ${\bf S}$ and ${\bf O}$ with length $4096$. ${\bf S}$ and ${\bf O}$ are then concatenated, and feed into one fully connected layer. The output of this fully connected layer is a feature vector $\bf M$ with length $1024$. For each image, we model all background objects to one FBE vector ${\bf B}$ with length $100$ using $0.9$ as the forgetting factor. ${\bf M}$ and ${\bf B}$ are then combined to one $1124$ dimension feature vector ${\bf R}$. ${\bf R}$ is then fed into three fully connected layers to calculate the final predicate classification results.

In training phase, we use the ImageNet \cite{deng2009imagenet} pre-trained model to initialize most of model parameters, and fine-tune the model using the training set of the VRD dataset. The training data include raw images and ground truth bounding boxes along with their class IDs. During the training process, we use $0.005$ as the learning rate, $0.9$ momentum, and $0.0005$ weight decay. We train the network for $40000$ iterations, and in every iteration we only input $1$ image into the model. In test phase, we input test images and the corresponding ground truth bounding boxes and their class IDs to the FOE-VRD for predicate classification. For relationship detection, the ground truth bounding boxes and the corresponding class IDs are generated by Faster RCNN algorithm. 

\begin{table*}[h]
	\caption{Experimental results of predicate classification task on the VRD test set, which include the results on entire test set (the second column) and the zero-shot test set (the third column). The measurement metric is Recall@100 (k = 70).}
	\label{predicateresults}
	\vskip 0.15in
	\begin{center}
		\begin{small}
			\begin{sc}
				\begin{tabular}{lcccr}
					\toprule
					Algorithm                   & Entire Set            &  Zero-shot Subset \\
					\midrule
					VRD \cite{lu2016visual}    &   84.34                  &    50.04                \\
					LKD \cite{yu2017visual}    &   86.97                  &    74.65                \\
					DSR \cite{liang2018visual} &   93.18                  &    79.81               \\
					Zoom-Net \cite{yin2018zoom} &  90.59                  &    N/A                \\
					NLG \cite{liao2019natural} &   92.73                  &    90.52                \\
					LSV \cite{jung2019improving} & 95.18                  &    83.49                \\
					\midrule 
					FOE-VRD without FBE       &    87.04                 &    82.83               \\
					FOE-VRD                   &    89.39                 &    96.05                \\
					\bottomrule
				\end{tabular}
			\end{sc}
		\end{small}
	\end{center}
	\vskip -0.1in
\end{table*}

For the relationship detection task, we use the VGG-16 \cite{simonyan2014very} based Faster-RCNN model (also implemented on TensorFlow 1.6 \cite{tensorflow2015-whitepaper} platform) to generate bounding boxes and their class IDs of the test images. Specifically, we use the well-trained VGG-16 model (using ImageNet \cite{deng2009imagenet}) to initialize the Faster-RCNN model, and then use the training set of the VRD dataset to finetune it. During the training phase of Faster RCNN, we use $0.001$ as the learning rate, $0.9$ momentum, and $0.0005$ weight decay. We use $256$ as the minibatch size, which means that every time we sample $256$ region proposals for training. We set one region proposal as positive sample if its $IOU > 0.7$ with some ground truth bounding boxes, and as negative sample if its $IOU < 0.3$. 

Our computation platform includes Intel Core i7 9700k CPU, 32 GB memory, and Nvidia GTX 1080 Ti GPU.

\subsection{Experimental Results}

In this part, we present the experimental results of the proposed FOE-VRD algorithm on the test set of VRD dataset. We compare FOE-VRD with several state-of-the-art baselines in case of predicate classification and relationship detection. Following \cite{yu2017visual}, we also introduce the hyperparameter $k$ into our experiments, and we set $k=70$ in our predicate classification and relationship detection tests. 

\begin{table*}[h]
	\caption{Experimental results of relationship detection task on the VRD test set, which include the results on entire test set (the second column) and the zero-shot subset (the third column). The measurement metric is Recall@100 (k = 70).}
	\label{relationshipresults}
	\vskip 0.15in
	\begin{center}
		\begin{small}
			\begin{sc}
				\begin{tabular}{lcccr}
					\toprule
					Algorithm                  & Entire Set            &  Zero-shot Subset  \\
					\midrule
					VRD \cite{lu2016visual}    &   21.51                  &    11.70                \\
					LKD \cite{yu2017visual}    &   31.89                  &    15.89                \\
					DSR \cite{liang2018visual} &   23.29                  &    9.20               \\
					DSL \cite{zhu2018deep}     &   18.26                  &    N/A                \\
					Zoom-Net \cite{yin2018zoom} &  27.30                  &    N/A                \\
					NLG \cite{liao2019natural}  &   21.97                 &    22.03                \\
					LSV \cite{jung2019improving} & 20.54                  &    12.14                \\
					\midrule 
					FOE-VRD without FBE       &    26.46                 &    23.54               \\
					FOE-VRD                   &    28.19                 &    25.91                \\
					\bottomrule
				\end{tabular}
			\end{sc}
		\end{small}
	\end{center}
	\vskip -0.1in
\end{table*}

\subsubsection{Predicate Classification}

Firstly, we evaluate FOE-VRD on predicate classification task. Table~\ref{predicateresults} presents the experimental results of predicate classification (include entire VRD test set and zero-shot subset) on several baselines and FOE-VRD. Notice that to show the importance of the proposed FBE feature, we remove FBE from the FOE-VRD model and use it as another baseline. From the experimental results we can learn that the introducing of FBE feature vector obviously improves the performance on both entire test set and zero-shot subset. Also, the proposed FOE-VRD algorithm has comparable or even better performance on predicate classification task comparing with the state-of-the-art methods, especially on zero-shot tests.

\subsubsection{Relationship Detection}

Relationship detection is a harder task than predicate classification, since it requires us firstly detect the possible objects in the images, then find potential relationships between them. In this task we firstly apply Faster-RCNN to do object detection and classification, then use FOE-VRD to recognize the predicate between every pair of detected objects. Table~\ref{relationshipresults} shows the experimental results of relationship detection (also include entire VRD test set and zero-shot subset). Like the results of predicate classification, the introducing of FBE feature vector also improves the performance of relationship detection. Moreover, on relationship detection, the proposed FOE-VRD algorithm has state-of-the-art performance, especially on zero-shot tests.

\section{Conclusion}
\label{Sec_Conclu}

In this paper, we propose a novel visual relationship detection algorithm named FOE-VRD. Motivated by the consideration that the classes of relationships are not only related to subjects and objects, but also background objects, we introduce Fixed-size Background Encoding (FBE) method, which successfully use fixed-size vectors to encode all background objects in images, despite the different number of them. Moreover, ROI Pooling method is also applied to model subjects and objects fixed-sizely. The fixed-size feature vectors make it possible to use fully-connected neural networks to do predicate classification. The experimental results on VRD dataset show that FOE-VRD works well on both predicate classification and relationship detection tasks. Moreover, FOE-VRD has good performance on zero-shot cases, which implies that it has good ability to deal with wide variety of real-world visual relationship detection tasks. 


%

\clearpage
%
%
\bibliographystyle{splncs04}
\bibliography{FOE_VRD}
\end{document}